\documentclass[runningheads]{llncs}
\usepackage[T1]{fontenc}
\usepackage{amsmath,graphicx,float, tikz, multirow, hyperref, listings, xcolor}

\lstdefinestyle{mystyle}{
    backgroundcolor=\color{lightgray!20}, % Background color
    commentstyle=\color{green},
    keywordstyle=\color{blue},
    numberstyle=\tiny\color{gray},
    stringstyle=\color{red},
    basicstyle=\ttfamily\small,
    breaklines=true, % Wrap long lines
    breakatwhitespace=true,
    numbers=left, % Line numbers on the left
    numbersep=5pt, % Space between line numbers and code
    frame=single, % Frame around the code
    captionpos=b, % Caption position
    tabsize=4, % Tab size
    showspaces=false, % Do not show spaces
    showstringspaces=false, % Do not show string spaces
    showtabs=false, % Do not show tabs
    keepspaces=true, % Keep spaces in text
    escapeinside={\%*}{*)}, % If you want to add LaTeX within your code
    columns=flexible, % Align code properly
}

\usepackage{graphicx}
\begin{document}
\title{Dynamic HumTrans: Humming Transcription Using CNNs and Dynamic Programming}
\author{Shubham Gupta\inst{1, 2} \and
Isaac Neri Gomez-Sarmiento\inst{2} \and
Faez Amjed Mezdari\inst{2} \and Mirco Ravanelli\inst{1, 3} \and Cem Subakan\inst{1, 2, 3}}
\authorrunning{Gupta et al.}
\institute{Mila-Québec AI Institute \and
Laval University \and Concordia University }

\titlerunning{Dynamic HumTrans: Humming Transcription}
\maketitle              % typeset the header of the contribution
\begin{abstract}
We propose a novel approach for humming transcription that combines a CNN-based architecture with a dynamic programming-based post-processing algorithm, utilizing the recently introduced HumTrans dataset. We identify and address inherent problems with the offset and onset ground truth provided by the dataset, offering heuristics to improve these annotations, resulting in a dataset with precise annotations that will aid future research. Additionally, we compare the transcription accuracy of our method against several others, demonstrating state-of-the-art (SOTA) results. All our code and corrected dataset is available at \url{https://github.com/shubham-gupta-30/humming_transcription}

\keywords{Humming \and Transcription \and Automatic Music Transcription (AMT) \and Music Information Retrieval (MIR)}
\end{abstract}
\section{Introduction}

The field of Automatic Music Transcription (AMT) has made significant progress in developing algorithms that transform acoustic music signals into music notation, positioning it as the musical analogue to Automatic Speech Recognition (ASR). In the piano-roll convention, a musical note is typically characterized by a constant pitch (related to frequency), onset time (the start), and offset time (the end).

\noindent One application of AMT is humming transcription, which involves extracting musical notes from a hummed tune. This is a crucial component for melody search engines \cite{googlehumming} and automatic music compositions \cite{basic_pitch_spotify}. Such applications enable song identification by mere humming, provide a quick starting point for creating new songs, and democratize music creation for those who may not play an instrument or have disabilities. However, achieving error-free music transcription remains a complex challenge, even for professionals.

\noindent In this paper, we explore humming transcription while working with the HUMTRANS dataset \cite{liu2023humtrans}, a novel dataset that claims to be the largest humming dataset to date. This dataset is a large collection of clean monophonic humming samples gathered by soliciting help from music students. This provides us with an opportunity for studying transcription in a monophonic setting. Various works in literature explore transcription in  more general polyphonic setting like VOCANO \cite{vocano}, Sheet Sage \cite{donahue2022melody}, MIR-ST500 \cite{MIR-ST500}, and JDC-STP \cite{kum2022pseudo}. We propose a novel approach to do an accurate transcription in this monophonic setting and show that we obtain state-of-the-art (SOTA) transcription results.  Our contributions are two fold:
\begin{enumerate}
    \item We identify issues in the ground truth provided by the HUMTRANS dataset and offer heuristics to address them, enabling us to bootstrap the creation of a high-quality subset with more meaningful annotations, which will aid further research in this direction.
    \item We introduce a novel approach that combines a CNN-based architecture with a dynamic programming-based post-processing technique, achieving state-of-the-art (SOTA) results.
\end{enumerate}

\subsection{Evaluation metrics}
The authors of the HumTrans dataset utilize the library $mir\_eval$ \cite{raffel2014mir_eval} to evaluate the performance of transcription methods on their dataset. The primary motivation for using this library is to standardize the implementation of metrics for music transcription. More specifically, they employ the method \\ \texttt{precision\_recall\_f1\_overlap}, which, according to the documentation, computes the Precision, Recall, and F-measure for reference vs. estimated notes. Correctness is determined based on note onset, pitch, and, optionally, offset, which the authors do not consider.

\noindent The authors consider a strict pitch tolerance of ±1 cent ($mir\_eval$ default value is ±50 cents), or in other words one hundredth part of a semitone, and the default onset tolerance of 50 ms. 

\noindent Precision, recall and F1-score are metrics that depend on the definition of true positives (TP), false negatives (FN) and false positives (FP) \cite{bock2012evaluating}.

\begin{equation}
    Precision=\frac{TP}{TP+FP}
\end{equation}

\begin{equation}
    Recall=\frac{TP}{TP+FN}
\end{equation}

\begin{equation}
    F1\, Score=2*\frac{Precision*Recall}{Precision + Recall}
\end{equation}

\noindent \textbf{TP}: Estimated onset is within the tolerance of the ground truth onset and is within the pitch tolerance.

\noindent \textbf{FP}: If N estimated onsets are within the tolerance of the ground truth, only 1 will be considered TP if it's also within the pitch tolerance and the rest N-1 estimated onsets will be considered FP. This means that all ground truth onsets can only be matched once.

\noindent \textbf{FN}: No estimated onsets were detected within the tolerance of the ground truth onset.

\subsection{Dataset Challenges}
A major issue with the HUMTRANS dataset is that the ground truth onsets and offsets are not well aligned. This can be explained because in their methodology they instructed their subjects to synchronize their humming with the rhythm of the played
melody, calling this approach as "self-labeling", without any post-processing. We had to overcome this challenge by coming up with semi-supervised ways to correct the provided onsets and offsets.

\subsection{Octave Aware vs Octave Invariance}
To simplify the problem of pitch estimation, the ground truth pitch given in MIDI file format can be transformed to an octave invariant representation by taking the modulo 12 of the MIDI numbers, which makes the song to be represented only by 12 semitones. In our work we produce results for both octave aware and octave invariant variations of the problem.

\section{Transcription methodology}

\subsection{Better ground truth annotation}
\label{sec:better_annotation}
For the purpose of training a neural network, we need precise onsets and offsets for ground truth. We realized that more accurate labeled onsets and offsets result in better-trained models. \\
To this end, we designed a heuristic-based algorithm to compute improved onsets and offsets. This algorithm calculates a {\it waveform envelope} and determines onsets and offsets based on when this envelope dips below a specific threshold value. The onsets and offsets obtained in this way can be noisy, so we refine them by eliminating spurious onsets/offsets, enforcing a minimum note length, and maintaining a minimum silence length between notes. This method is supervised by using the number of notes from the ground truth provided by the dataset. We retain only those training, testing, and validation samples where the number of notes detected by our heuristic matches the number provided by the ground truth.\\
It's important to note that we can only trust the number of notes from the ground truth, as the onsets and offsets cannot be relied upon. By using this approach, we obtained better ground truth onsets and offsets, retaining $6,827$ of the $13,080$ in the training set ($52.2\%$) and $440$ of the $769$ (i.e., $57\%$) in the test set. More details on this method are covered in Appendix \ref{heuristic}.

\begin{figure}[ht]
  \centering
  \includegraphics[width=0.9\textwidth,height=0.25\textheight,keepaspectratio]{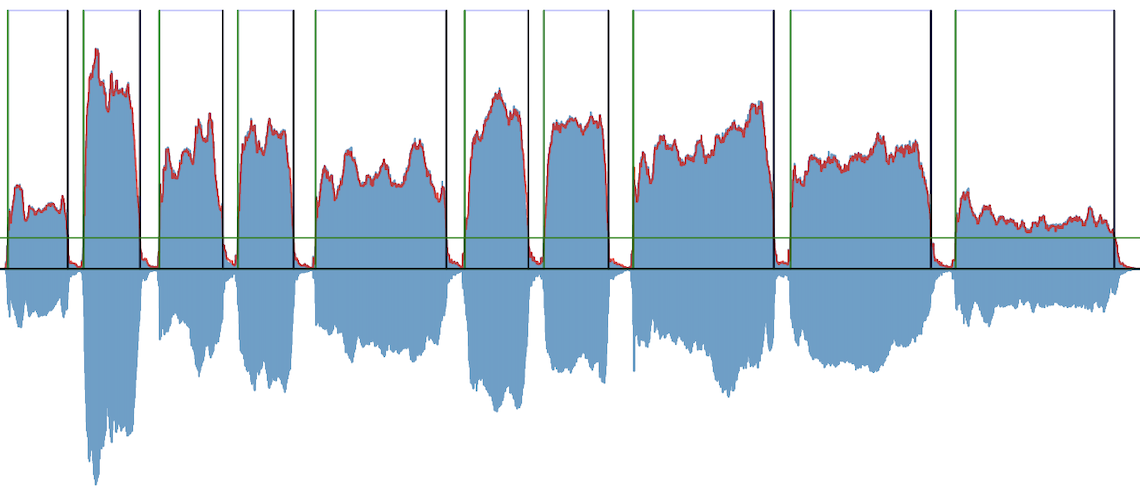}
  \includegraphics[width=0.9\textwidth,height=0.25\textheight,keepaspectratio]{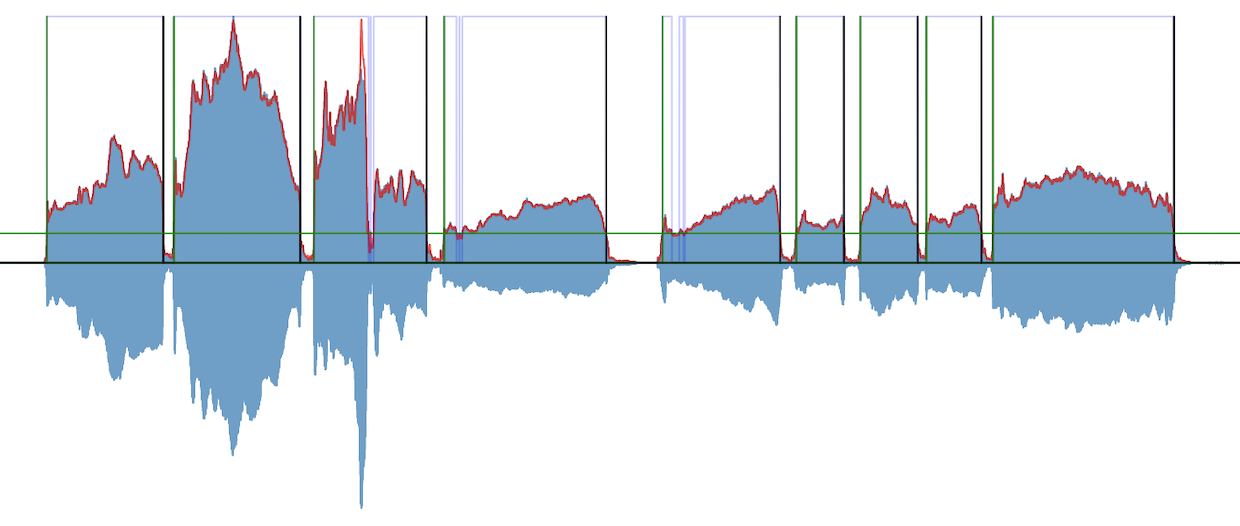}
  \caption{Examples where the heuristic algorithm for onset and offset detection succeeds (top) and fails (bottom).}
  \label{fig:onset_offset_algorithm}
\end{figure}

\begin{figure}[ht]
  \centering
  \includegraphics[width=0.8\textwidth,height=0.5\textheight,keepaspectratio]{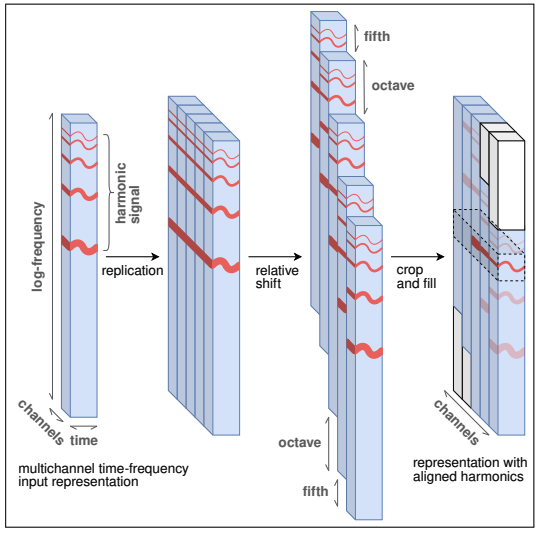}
  \caption{Harmonic Stacking. Source: \cite{balhar10melody}}
  \label{fig:harmonic_stacking}
\end{figure}

\subsection{Network Design}
Our neural network architecture is a convolution-based network. The model is inspired by the architecture in \cite{bittner2022lightweight}. We have tailored our network for our use case of a monophonic humming dataset. The following are key elements of this design:
\begin{enumerate}
    \item {\bf Input representation}: While $STFT$ and Mel spectrograms work well for speech-related tasks, $CQT$ representation is much more effective for $MIR$ (Music Information Retrieval) as the geometric structure of this transform closely matches the geometric nature of Western classical music.
    \item {\bf Harmonic Stacking}: While instruments can produce desired notes precisely, humans aren't very adept at doing so. Human singing/humming normally includes not only the main note but also overtones, which can be spatially far apart from each other in a CQT transform. To address this, \cite{balhar10melody} introduces Harmonic Stacking, where CQT time frames are shifted by the right amount of offsets to bring overtones closer to each other
\end{enumerate}

% \tikzstyle{layer} = [rectangle, rounded corners, minimum width=3cm, minimum height=1cm,text centered, draw=black, fill=blue!30]
% \tikzstyle{specialLayer} = [rectangle, rounded corners, minimum width=2cm, minimum height=0.5cm, text centered, draw=black, fill=orange!30]

% \tikzstyle{arrow} = [thick,->,>=stealth]

% \begin{tikzpicture}[node distance=1.3cm]
% % Title
% \node [align=center] at (0,1) {\bf Model Structure};

% \node (input) [specialLayer] {Input Layer (wav\_data)};
% \node (transforms) [layer, below of=input] {CQT Transform + Harmonic Stacking};
% \node (conv1) [layer, below of=transforms, align=center] {Conv2D(16 channels, (5, 5) kernel) \\ + BatchNorm + ReLU + Dropout};
% \node (conv2) [layer, below of=conv1, align=center] {Conv2D(8 channels, (3, 39) kernel) \\  + BatchNorm + ReLU + Dropout};
% \node (conv3) [layer, below of=conv2, align=center] {Conv2D(32 channels, (7, 7) kernel) \\ + BatchNorm + ReLU + Dropout};
% \node (conv4) [layer, below of=conv3, align=center] {Conv2D(1 channel, (7, 3) kernel)};
% \node (final) [layer, below of=conv4] {Final Linear Layer};
% \node (output) [specialLayer, below of=final] {Output - Affinity matrix};

% \draw [arrow] (input) -- (transforms);
% \draw [arrow] (transforms) -- (conv1);
% \draw [arrow] (conv1) -- (conv2);
% \draw [arrow] (conv2) -- (conv3);
% \draw [arrow] (conv3) -- (conv4);
% \draw [arrow] (conv4) -- (final);
% \draw [arrow] (final) -- (output);

% \end{tikzpicture}

\begin{figure}[ht]
  \centering
  \includegraphics[width=0.3\textwidth]{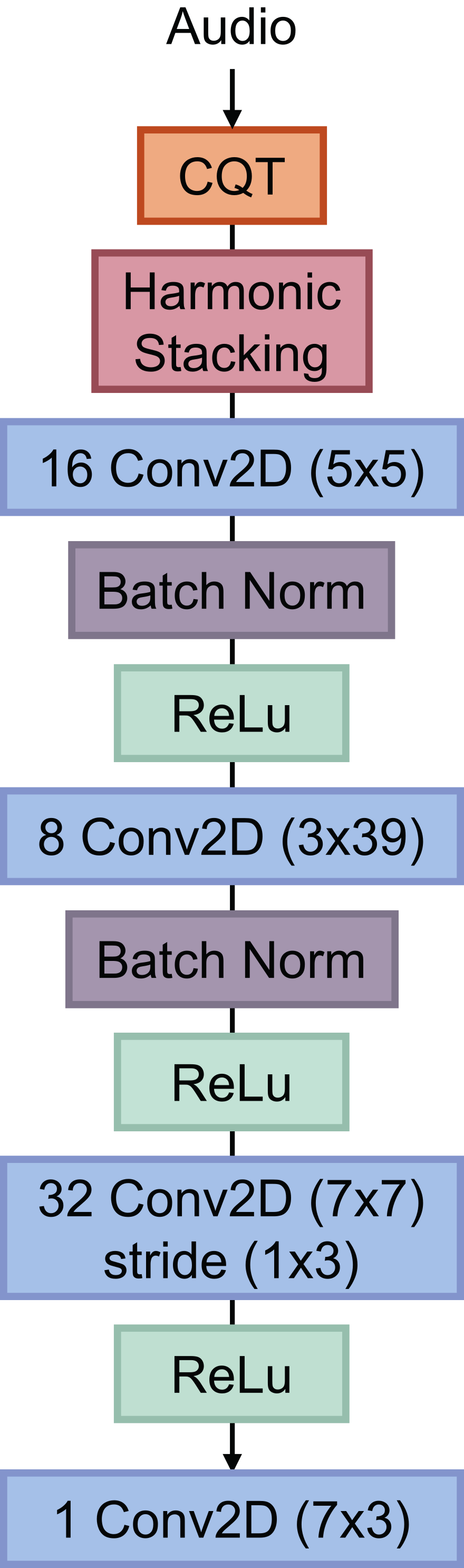}
  \caption{Our model architecture - A minimal version of Spotify's BasicPitch model.}
\end{figure}

\subsection{Training}
We train the model with a batch size of $16$. For each batch, we randomly select a small sample from the humming recordings by choosing $5$ to $10$ notes for each batch element. Additionally, we introduce a new dummy note $89$  to signify the beginning and end of the recording sample, as well as the silence between notes. Our task for every time frame of the CQT representation is to predict the note that the frame represents. The model is trained using {\bf CrossEntropy} loss and employs {\bf Adam} as the optimizer with a learning rate of $0.001$. Unlike other works in this field that first predict onsets and offsets and then condition note prediction on them, as in \cite{basic_pitch_spotify}, our method infers onsets and offsets directly from the predicted notes.

\subsection{Inference}
\label{inference}
Unlike transformers, convolution networks generalize well beyond the training length examples they are trained on. Because of this, we do not have to perform inference on pieces of fixed lengths; instead, we can perform inference on the entire sample as long as available memory allows. During inference, we calculate model logits for each time frame over the possible space of notes. The naive way to convert these logits to actual note predictions would be to take the note at each time frame with the maximum probability. However, this results in noisy note attributions. \\

\noindent We clean up these noisy attributions using a dynamic programming based algorithm inspired by the use of the {\it Viterbi algorithm} in text-to-speech alignment in speech tasks. Given the affinity matrix denoting the affinity scores of each time frame with the possible notes, we can define a path $P$ through this matrix as valid if it satisfies the following constraints (Note that the dummy note introduced is 89, and $T$ is the length of the segment being inferred):
\begin{enumerate}
    \item $P$ starts at $(0, 89)$ and ends at $(T-1, 89)$.
    \item If $P$ is at note $n \neq 89$ at time $t$, then at time $t+1$, it can be at either $n$ or $89$.
    \item If $P$ is at note $89$ at time $t$, then it can be at any note at time $t+1$.
\end{enumerate}
These constraints ensure that we do not switch abruptly from one note to another without going through the dummy note, which is a realistic constraint as in all humming samples, the space between two hummed notes is very noticeable. Using a dynamic programming-based method, we can find the path $P$ with the highest probability among all valid paths. We further clean up this path $P$ to enforce minimum note length constraints and report this final cleaned note assignment as our inferred note assignment. We read the onsets and offsets from these note assignments. We reproduce the code for this dynamic programming based postprocessing in Appendix \ref{app:inference}
\begin{figure}
  \centering
  \includegraphics[width=0.9\textwidth]{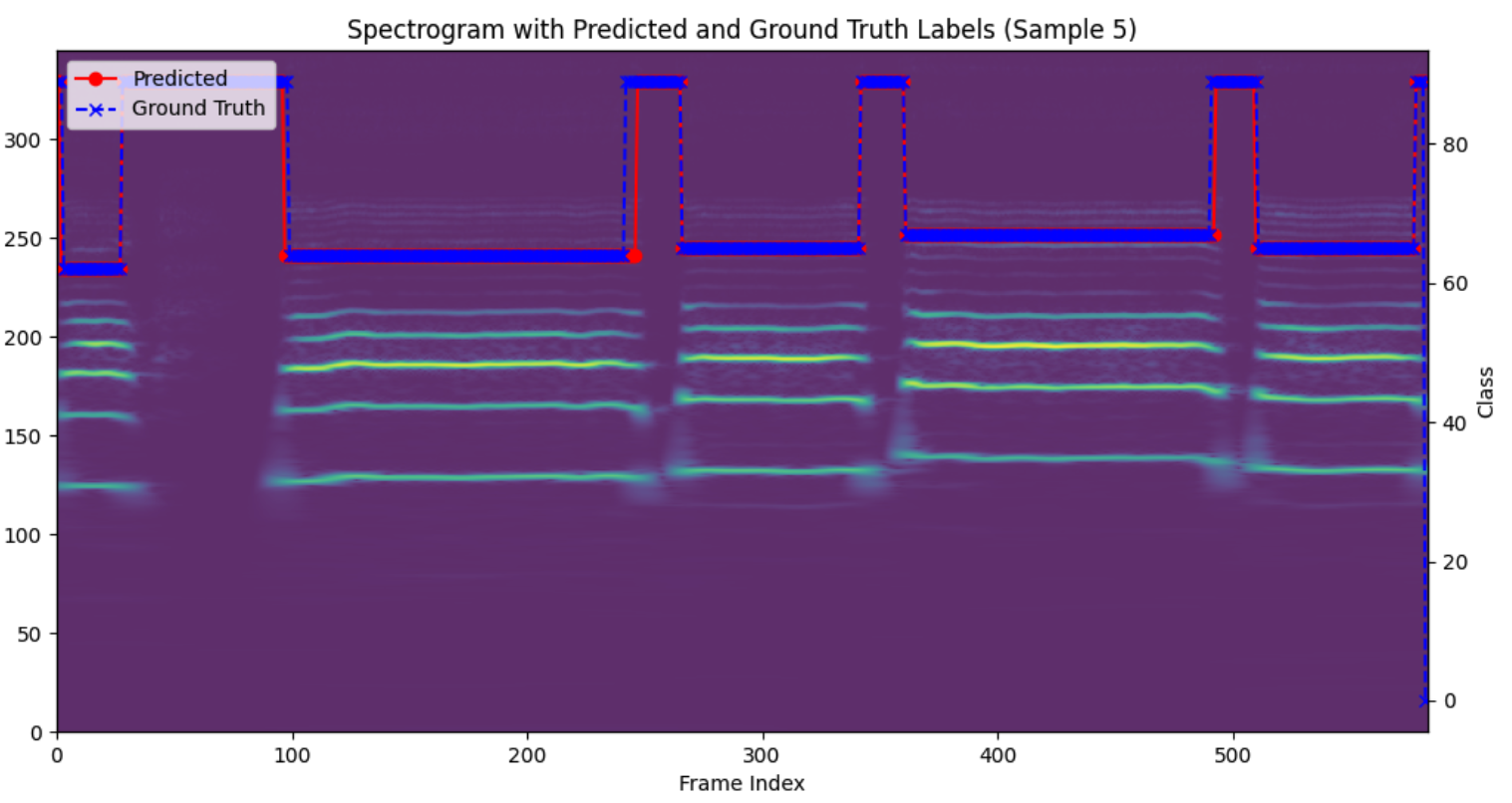}
  \caption{Example inference: blue represents the ground truth and red the inferred cleaned notes.}
\end{figure}

\section{Results and discussion}
\label{sec:results_discussion}
We compare our methods with various methods discussed in \cite{liu2023humtrans}. The authors provide extracted MIDIs for the test set for 4 methods in their GitHub repository \cite{humtrans_github}. These are - $Vocano$ \cite{vocano}, $MIR-ST500$ \cite{MIR-ST500}, $SheetSage$ \cite{donahue2022melody}, and $JDC-STP$ \cite{kum2022pseudo}. In addition, we compute MIDIs for the test set using Spotify's $basic\_pitch$ \cite{basic_pitch_spotify}, and compare these with the methods proposed in this report.

\subsection{Octave invariant}
Following \cite{liu2023humtrans}, we calculate precision, recall, and F1-score, using the $mir\_eval$ library, with an onset tolerance of $50ms$ and disregarding offsets. We provide two comparisons here - a comparison with respect to the corrected ground truth we obtain and a comparison where we measure only the note accuracy, disregarding both onsets and offsets. Additionally, these comparisons are octave invariant, i.e a note is considered to be correctly predicted even if the octave does not match the ground truth exactly. We provide these results on the test set provided by the dataset in table \ref{tab:octave_invariant}

\begin{table}[ht]
\label{tab:octave_invariant}
\centering
\resizebox{\columnwidth}{!}{%
\begin{tabular}{|c||c|c|c||c|c|c|}
\hline
\multirow{2}{*}{Method} & \multicolumn{3}{c||}{Note + Onsets} & \multicolumn{3}{c|}{Notes Only} \\
\cline{2-7}
& P & R & F1 & P & R & F1 \\
\hline

{\bf Ours} & {\bf 0.670} & {\bf 0.675} & {\bf 0.673} & {\bf 0.848} & \bf{0.854} & \bf{0.850} \\
\hline
{VOCANO} & 0.568 & 0.561 &  0.564 & 0.729 & 0.723 &  0.726 \\
\hline
{JDC-STP} & 0.502 & 0.487 & 0.490 & 0.795 & 0.784 & 0.783 \\
\hline
{SheetSage} & 0.171 & 0.170 & 0.170 & 0.446 & 0.442 & 0.444 \\
\hline
{MIR-ST500} & 0.601 & 0.608 & 0.604 & 0.808 & 0.820 & 0.813 \\
\hline
{basic\_pitch} & 0.392 & 0.497 & 0.434 & 0.653 & 0.847 & 0.729 \\
\hline
\end{tabular}
}
\caption{Octave Invariant metrics computed on test set.}
\end{table}

\subsection{Octave aware}
In table \ref{tab:octave_aware}, we also provide comparisons of our method with other methods while requiring the models to be octave-aware, i.e., we are looking for an exact note match, including the correct octaves.

\begin{table}
\label{tab:octave_aware}
\centering
\resizebox{\columnwidth}{!}{%
\begin{tabular}{|c||c|c|c||c|c|c|}
\hline
\multirow{2}{*}{Method} & \multicolumn{3}{c||}{Note + Onsets} & \multicolumn{3}{c|}{Notes Only} \\
\cline{2-7}
& P & R & F1 & P & R & F1 \\
\hline
{\bf Ours} & {\bf 0.649} & {\bf 0.653} & {\bf 0.651} & {\bf 0.814} & \bf{0.820} & \bf{0.817} \\
\hline
{VOCANO} & 0.344 & 0.340 &  0.341 & 0.446 & 0.443 &  0.444 \\
\hline
{JDC-STP} & 0.297 & 0.279 & 0.286 & 0.463 & 0.442 & 0.450 \\
\hline
{SheetSage} & 0.161 & 0.160 & 0.161 & 0.434 & 0.430 & 0.444 \\
\hline
{MIR-ST500} & 0.360 & 0.363 & 0.361 & 0.486 & 0.491 & 0.488 \\
\hline
{basic\_pitch} & 0.243 & 0.304 & 0.268 & 0.388 & 0.498 & 0.432 \\
\hline
\end{tabular}
}

\caption{Octave Aware metrics computed on test set.}
\end{table}

\subsection{Discussion}
\noindent We observe that our method outperform all tracked methods for humming transcription.   We observe that SheetSage performs the worst in all comparisons. Also note that our method performs similarly well in the octave invariant and the octave aware setting, indicating that our architecture is able to learn very robust note representations.

\subsection{Future Work}
In our work, we provide a novel methodology to accurately estimate monophonic humming transcriptions. A natural extension of this work is to transcribe polyphonic humming samples. This work also provides a novel dynamic programming based post processing and we would like to explore the use of this as postprocessing in other transcription problems. It is also possible to use this postprocessing as a part of the loss function during training thus enabling better transcriptions from the get go. 
\bibliographystyle{splncs04}
\bibliography{paper}

\appendix
\clearpage

\section{Heuristic Algorithm for Better Ground Truth Annotations}
\label{heuristic}
We reproduce the python code used to create the waveform envelopes as mentioned in Section \ref{sec:better_annotation}. We want our waveform envelops to be as close to the original general waveform shape as possible and thus we utilize a simple heuristic algorithm that locally computes the maximum of the waveform preceding a point of interest and the maximum of the waveform following a point of interest. We found we get a very tight hugging envelope when we take a min of these two values. 

\begin{lstlisting}[language=Python, caption=Calculate waveform envelope, label=lst:waveform]
def get_waveform_envelope(signal):

    # Padding for the signal
    padding = 100
    padded_signal = torch.nn.functional.pad(
        signal,
        (padding, padding),
        mode='constant',
        value=0)

    # Unfold to get sliding windows
    windows_before = padded_signal[:-padding].unfold(0, padding, 1)
    windows_after = padded_signal[padding:].unfold(0, padding, 1)

    # Compute maximums
    max_before = windows_before.max(dim=1).values
    max_after = windows_after.max(dim=1).values

    # Compute minimum of the two maximums
    modified_signal = torch.min(max_before, max_after)

    return modified_signal
\end{lstlisting}

We found that the envelope calculated using the above method could still be improved if we calculated the envelope of the envelope again. Having now obtained a tight envelope of the waveform, we now calculate the threshold to use for this waveform to measure the onset and offset boundaries. We further clean these onsets and offsets by disregarding any silences that are too small (the method \texttt{adjust\_onsets\_offsets} in the code below). Finally, we check if only consider this waveform for training or testing purposes if we get the right number of notes through this heuristic, otherwise we disregard this sample altogether. The code to do this is reproduced below. 
\clearpage
\begin{lstlisting}[language=Python, caption=Process a single waveform, label=lst:process]
    def process(i):    
        signal =  torch.abs(torch.tensor(dataset[i]["wav_data"]).float())
        envelope = get_waveform_envelope(signal)[None]
        envelope = get_waveform_envelope(envelope)[None]
        mw_min = torch.min(envelope)
        mw_max = torch.max(envelope)
        thresholds_for_i = mw_min + thresholds * (mw_max - mw_min)
        above_threshold = envelope > thresholds_for_i
    
        num_notes_known = dataset[i]["midi_notes"].shape[0]
    
    
        for t_idx in range(len(threshold_values)):
            current_above_threshold = above_threshold[t_idx]
            onsets = (current_above_threshold[:-1] < current_above_threshold[1:]).nonzero(as_tuple=True)[0]
            offsets = (current_above_threshold[:-1] > current_above_threshold[1:]).nonzero(as_tuple=True)[0] + 1
    
            onsets, offsets = adjust_onsets_offsets(onsets, offsets, envelope.shape[0])
            num_notes_discovered = len(offsets)
            if num_notes_discovered == num_notes_known:
                    file_path = filtered_folder / f"{dataset[i]['file_name']}_onsets_offsets.txt"
                    with file_path.open('w') as f:
                        for onset, offset in zip(onsets, offsets):
                            f.write(f"{onset} {offset}\n")
                    plot_waveforms(
                        signal.numpy(),
                        envelope.numpy(),
                        thresholds_for_i[0].item(),
                        current_above_threshold.numpy(), onsets, offsets, f"{i}_{dataset[i]['file_name']}.png")
                    return True
    
        return False
\end{lstlisting}
\clearpage
\section{Dynamic Programming Postprocessing}
\label{app:inference}
The code for computing a path using dynamic programming as detailed in section \ref{inference} is reproduced below. Note that the method \texttt{clean\_path} simply performs a heuristic cleaning on the paths discovered by the dynamic programming solution so that they are more meaningful and make sense. 

\begin{lstlisting}[language=Python, caption=Dynammic Programming Postprocessing, label=lst:process2]
    def build_log_path_prob_matrix_with_path(log_affinity):
        L, T = log_affinity.shape
        prob_table = np.full((L, T), float('-inf'))
        path_matrix = np.zeros((L, T, 2), dtype=int)
    
        # Base case
        prob_table[L-1, 0] = 0
    
        # Iterating over columns
        for j in range(1, T):
            # Vectorized computation for non-last rows
            non_last_rows = np.arange(L-1)
            max_vals = np.maximum(prob_table[non_last_rows, j-1], prob_table[L-1, j-1])
            prob_table[non_last_rows, j] = max_vals + log_affinity[non_last_rows, j]
            path_matrix[non_last_rows, j] = np.vstack([non_last_rows, np.full(L-1, j-1)]).T
            path_matrix[non_last_rows, j, 0] = np.where(prob_table[non_last_rows, j-1] == max_vals, non_last_rows, L-1)
    
            # Computation for last row, still iterative
            max_index = np.argmax(prob_table[:, j-1])
            prob_table[L-1, j] = prob_table[max_index, j-1] + log_affinity[L-1, j]
            path_matrix[L-1, j] = (max_index, j-1)
    
        # Retrace path
        path = []
        current_pos = (L-1, T-1)
        while current_pos[1] != 0:
            path.append(current_pos)
            current_pos = tuple(path_matrix[current_pos])
    
        path.append(current_pos)
        path = path[::-1]
        path = clean_path(path, log_affinity, L-1)
        return prob_table, path

\end{lstlisting}

\end{document}